\documentclass{article}

\usepackage[preprint]{neurips_2023}

\usepackage[utf8]{inputenc} 
\usepackage[T1]{fontenc}    
\usepackage{hyperref}       
\usepackage{url}            
\usepackage{booktabs}       
\usepackage{amsfonts}       
\usepackage{nicefrac}       
\usepackage{microtype}      
\usepackage{xcolor}         
\usepackage{multirow,adjustbox}

\usepackage[]{color-edits}
\addauthor{ch}{red}
\addauthor{by}{blue}
\makeatletter
\def\blfootnote{\gdef\@thefnmark{}\@footnotetext}
\makeatother

\title{Instance-Aware Repeat Factor Sampling for Long-Tailed Object Detection}

\author{%
  Burhaneddin Yaman \\
  Bosch Research North America\\
  Bosch Center for AI\\
  \texttt{burhaneddin.yaman@us.bosch.com} \\
  \And
  Tanvir Mahmud \\
   University of Texas at Austin \\
  \texttt{tanvirmahmud@utexas.edu} \\
  \AND
  Chun-Hao Liu \\
  Amazon Prime Video \\
  \texttt{chunhaol@amazon.com} \\
}

\begin{document}

\maketitle

\begin{abstract}
We propose an embarrassingly simple method -- instance-aware repeat factor sampling (IRFS) to address the problem of imbalanced data in long-tailed object detection. Imbalanced datasets in real-world object detection often suffer from a large disparity in the number of instances for each class. To improve the generalization performance of object detection models on rare classes, various data sampling techniques have been proposed. Repeat factor sampling (RFS) has shown promise due to its simplicity and effectiveness. Despite its efficiency, RFS completely neglects the instance counts and solely relies on the image count during re-sampling process. However, instance count may immensely vary for different classes with similar image counts. Such variation highlights the importance of both image and instance for addressing the long-tail distributions. Thus, we propose IRFS which unifies instance and image counts for the re-sampling process to be aware of different perspectives of the imbalance in long-tailed datasets. Our method shows promising results on the challenging LVIS v1.0 benchmark dataset over various architectures and backbones, demonstrating their effectiveness in improving the performance of object detection models on rare classes with a relative $+50\%$ average precision (AP) improvement over counterpart RFS. IRFS can serve as a strong baseline and be easily incorporated into existing long-tailed frameworks. 
\end{abstract}

\section{Introduction}
\blfootnote{Work done in Bosch Research North America.}
 Real-world datasets are often imbalanced and follow a long-tail distribution where few object classes dominates the distribution resulting in the heavy tail for rare classes~\cite{liu2019large,lvis,zhao2022fastmri+, liu2022object}. This poses a significant challenge in many real-world applications, especially in computer vision tasks such as image recognition, object detection, and semantic segmentation~\cite{lvis}. Without addressing this issue, models tend to be biased towards the majority classes, making it difficult to achieve good performance in the rare classes~\cite{Tan_2020_CVPR,Chang2021ImageLevelOO,alexandridis2022long,Hu_2022_CVPR}. However, performing well on rare classes is crucial for applications like autonomous driving, video surveillance, automated optical inspection, and medical imaging systems, where anomalies or unusual events can have significant implications. Therefore, addressing the long-tail distribution of real-world visual data is crucial for achieving reliable performance in many real-world applications.

The problem of data imbalance with a long-tail distribution has been approached in two major directions. The first direction involves data augmentation with class-balanced sampling~\cite{lvis,Chang2021ImageLevelOO,zhang2021mosaicos}. While up-sampling the minor classes through data augmentation and down-sampling the major classes may appear promising, such an approach may lead to performance bias towards the largely augmented minor classes. The utilization of object-centric sampling and class-balanced sampling can mitigate this issue by exploiting available data and achieving better performance on minor classes. However, significant imbalance between major and minor classes may require much larger up-sampling and down-sampling ratios that under-represent major classes and make training biased. The second direction involves modifying the loss function to handle the effect of large class imbalance~\cite{li2020overcoming,Tan_2021_CVPR,alexandridis2022long}. Adaptive weighting of the classes in the loss function is one such solution that adds more weight to the minor class predictions. Grouping the predictions and loss functions according to the available samples of each class is another promising solution. Despite considerable improvement of performance with these approaches, the larger imbalance of the dataset can significantly under-represent the minor classes that suffer from similar challenges. 

\begin{figure}[t]
\centering
\includegraphics[width=\linewidth]{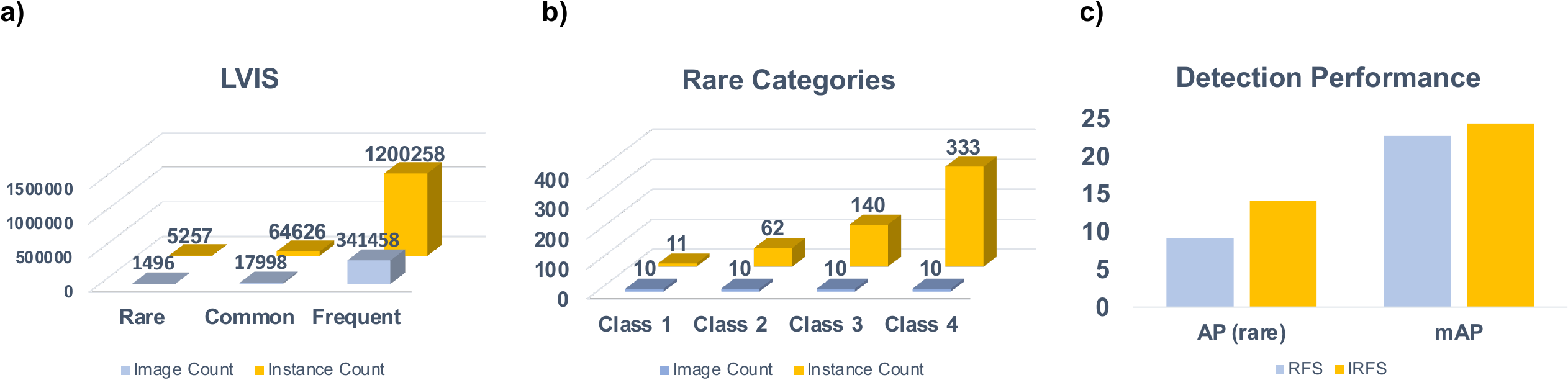}
\caption{We illustrate the importance of instance count in addressing long-tail distributions. Figure a) shows image and instance counts for the LVIS v1.0 data distribution across three main categories, namely rare, common, and frequent. Figure b) demonstrates a representative case from the rare categories with the same image count but varying instance counts. It is observed that RFS, which ignores the instance count, assigns the same re-sampling factor to all classes with the same image count. In contrast, figure c) depicts the IRFS approach, which incorporates the instance count in the re-sampling process, resulting in a $+50\%$ relative improvement compared to RFS on rare categories and a better overall performance.}
\label{fig:IRFS_fig}
\end{figure}

While class-based sampling approaches have focused on the imbalance in terms of the number of images in the dataset, instance level imbalance remains unaddressed. As a result, current class balanced sampling approaches assign the same oversampling ratio to all categories with the same number of images, despite the immense amount of imbalance in terms of instances (see Fig.~\ref{fig:IRFS_fig}). Thus, long-tailed datasets are not only imbalanced in terms of images but also instances. Therefore, disregarding instance count while addressing the long-tail distributions may be costly and lead to suboptimal results, especially for rare categories. To address this issue, we propose a novel instance-aware repeat factor sampling (IRFS) method that incorporates the instance count and fuses it with image count to determine the repeat factor for each category. Our main contributions are threefold.
\begin{itemize}
    \item We propose an incredibly simple method to address long-tail distributions by incorporating the instance count in the re-sampling process.
    \item Our proposed approach can be used as a standalone replacement for current repeat factor sampling (RFS) approaches or combined with other re-weighting approaches.
    \item We conduct extensive experiments on challenging LVIS v1.0 benchmark. Our proposed instance-aware re-sampling approach provides significant performance improvement over RFS, and it achieves the state-of-the-art performance in both overall accuracy and accuracy of rare classes.
\end{itemize}

\section{Related Works}
\textbf{Object Detection and Instance Segmentation.}
The task of object detection has been an active research area in computer vision. In recent years, several approaches have been proposed to improve the accuracy and speed of object detection. The object detection models can be categorized into two main types: anchor-based~\cite{7410526,7780460} and anchor-free~\cite{Law_2018_ECCV,Duan_2019_ICCV} approaches. Anchor-based approaches utilize a set of pre-defined anchors to match predicted bounding boxes with ground truth bounding boxes. After matching, the model classifies the prediction and rectifies the offset for the box coordinates through regression. On the other hand, anchor-free approaches directly predict a set of key points for each object without prior anchors. These predicted key points are then grouped to determine the final bounding boxes. Furthermore, object detection models can be designed as either single-stage~\cite{10.1007/978-3-319-46448-0_2,7780460,8417976,9156454,zhang2020bridging} or two-stage models~\cite{7410526,10.5555/2969239.2969250,8237584,10.1007/978-3-030-58555-6_16}. Single-stage models predict bounding boxes directly with one network, making them well-suited for real-time applications due to their fast inference speed. In contrast, two-stage models first predict bounding box proposals and then perform box regression and classification on top of them. Finally, instance segmentation is a more fine-grained object detection task that not only detects the bounding boxes but also predicts the per-pixel class category within each box. Several classical instance segmentation models include Mask R-CNN~\cite{8237584}, Cascade R-CNN~\cite{Cai_2018_CVPR}, DetectorRS~\cite{9578795}, and recent Transformer-based detectors~\cite{DBLP:conf/cvpr/ChengMSKG22,li2022mask,chen2023vision}.

\textbf{Long-Tailed Challenges.} In the realm of computer vision, long-tail distribution poses a persistent challenge that has yet to be fully resolved~\cite{DBLP:journals/ijcv/YangJSG22}. Such distribution patterns are common in real-world datasets, wherein a small subset of categories dominate the majority of instances, and each category's importance does not necessarily follow a power-law decayed distribution. Achieving balanced performance for each category in this highly skewed distribution remains an open problem, particularly in the context of object detection, which is a more complex task than image recognition due to the presence of multiple classes in a single image. Various methods have been proposed to address long-tailed image classification, including assigning balanced weights for each class in the loss functions or adjusting the sampling rate for each class in the data loader~\cite{cui2019classbalancedloss,Kim2020M2mIC,hong2021disentangling,Alshammari_2022_CVPR}. However, applying these solutions to object detection is more challenging due to the nature of the task.

\textbf{Re-sampling in Long-Tailed Detection.} The state-of-the-art re-sampling technique in long-tailed detection -- repeat factor sampling, was introduced in~\cite{lvis}. This technique involves increasing the sampling rate of images that contain rare class objects, thereby improving the balance of the dataset. While RFS has proven to be effective, it fails to consider the object numbers factor jointly. In response to this limitation, the authors proposed re-sampling at object-level (OCS)~\cite{Chang2021ImageLevelOO}, which leverages a memory bank to replay the necessary objects in an image. The authors in~\cite{Chang2021ImageLevelOO} further proposed using RFS together with OCS to achieve a balanced sampling in both image and object levels. We note that this approach does not modify or integrate instance count into the image re-sampling process. Instead, it utilizes the features of rare categories from the memory bank when computing the loss. Applying data augmentation strategies is another type of re-sampling technique. It has also shown promising results in addressing the long-tailed problem in object detection. For example, simple copy-and-paste techniques~\cite{DBLP:conf/cvpr/GhiasiCSQLCLZ21} have been used to randomly mix objects from different images, leading to improved performance in long-tailed instance segmentation tasks. Similarly, another approach called MOSAICOS~\cite{zhang2021mosaicos} has demonstrated the effectiveness of random mixing multiple object-centric images into a single image, which can significantly boost the baseline performance of long-tailed object detection. 

\textbf{Re-weighting in Long-Tailed Detection.} Recently, the research community has shifted its focus towards inventing new re-weighting techniques to tackle the long-tailed detection problem. Equalization loss (EQL)~\cite{Tan_2020_CVPR,Tan_2021_CVPR} was proposed to alleviate discouraging gradients for rare categories during parameter updating, thus improving the learning of better features for rare classes. Li et al.~\cite{li2020overcoming} proposed a balanced group softmax (BAGS) module that formulates a novel type of group-wise training. This module separates the tail and head classes, thus improving the learning of similar numbers of instances within each group. Observing that the classifier’s weight norm can cause imbalance and make long-tailed detection challenging, the authors in~\cite{Wang_2022_CVPR} proposed a C2AM loss, which enforces the margin between two categories to be proportional to the ratio of their classifiers’ weight norms. Cho et al.~\cite{10.1007/978-3-031-20074-8_40} took a theoretical approach to derive a bound for object detection performance metric using classical margin-based binary classification theory. They proposed a surrogate objective named effective class-margin (ECM) loss. Specifically for instance segmentation, seesaw loss~\cite{DBLP:conf/cvpr/WangZZCPGCLLL21} and Gumbel optimized loss~\cite{alexandridis2022long} were proposed. The former dynamically re-balances gradients of positive and negative samples for each category, while the latter better aligns the distribution of the activation function with the long-tail distribution.

\section{Methods}
\subsection{Repeat Factor Sampling}
In this section, we introduce the Repeat Factor Sampling (RFS) method for training object detection models. RFS is a simple and effective approach that balances the class distribution by oversampling images containing rare classes~\cite{lvis}. Specifically, RFS increases the occurrence rate of tail categories by repeating the images that contain them during training. This method has shown to be effective in addressing long-tail distribution challenges in object detection and has demonstrated promising results in improving the performance of models on rare categories.

For each category $c$, let $f_c$ be the fraction of training images that contain at
least one instance of $c$. Accordingly, a category-level repeat factor $r_c$ is defined as
\begin{equation}\label{rfs}
r_c = \max(1, \sqrt{t/f_c}), 
\end{equation}
where $t$ is a hyper-parameter that intuitively controls the point at which oversampling begins. If $f_c$ is greater than or equal to $t$, then there is no oversampling for that category. Additionally, as each image may contain different categories, an image-level repeat factor $r_i$ is defined for each image $i$, which is determined as
\begin{equation}\label{rfs_image_level}
r_i = \max_{c\in i} r_c,
\end{equation}
where $c\in i$ denotes the categories labeled in the $i$-th image. During training, each image is repeated according to its repeat factor $r_i$.

\subsection{Proposed Instance-Aware Repeat Factor Sampling}
Class-balanced sampling through RFS has significantly improved performance, but exclusion of instances can lead to suboptimal results, especially for rare classes.  Figure \ref{fig:IRFS_fig}a) shows the image count and instance count for the three main categories. In addition to imbalance in image count, there exists a visibly significant further imbalance in terms of instance counts. Despite the varying instance count (Figure \ref{fig:IRFS_fig}b)), classes with the same image count are assigned the same re-sampling factor which might lead to suboptimal results (Figure \ref{fig:IRFS_fig}c)). Therefore, the $f_c$ parameter in RFS may not be representative enough to handle long-tail distributions, especially if there is an additional imbalance layer between the number of images and the number of bounding boxes.

In order to address the challenges faced by RFS and to ensure that rare classes are properly prioritized during the sampling process, we propose an instance-aware repeat factor sampling (IRFS) method that takes both images and bounding box instances into account. To be more specific, we propose using both the image and bounding box instances to define $f_c$. Let $f_{(i,c)}$ and $f_{(b,c)}$ denote the fractions of images and bounding boxes in the training set that contain instances of category $c$. We define $f_c$ as the mean of $f_{(i,c)}$ and $f_{(b,c)}$. For a geometric mean case, the Eq.~\ref{rfs} can be reformulated as 
\begin{equation}\label{irfs}
r_c = \max(1, \sqrt{t/\sqrt{f_{(i,c)} \times f_{(b,c)} }}).
\end{equation}
Once instance-aware category-level repeat factor is calculated in Eq.~\ref{irfs}, Eq. \ref{rfs_image_level} is used to compute the image-level repeat factor.
\begin{table}[t]
\caption{LVIS v1.0 validation set results using Mask R-CNN and Cascade Mask R-CNN frameworks with ResNet-50/ResNet-101 backbones on $1\times$ schedule. RFS and IRFS are trained with $t=10^{-3}$ and w/o sampling is trained with $t=0$.}
\label{tab:table_mask_rcnns}
\begin{adjustbox}{width=1.0\linewidth,center}
\begin{tabular}{lllllllll}
\toprule
Framework                   & Backbone                   & \textbf{Detection}      & mAP$_{bbox}$ & AP$_{50}$ & AP$_{75}$ & AP$_r$ & AP$_c$ & AP$_f$ \\ \midrule
\multirow{3}{*}{Mask R-CNN} & \multirow{3}{*}{ResNet-50} & w/o sampling      & 16.9      & 28.1    & 17.7          & 0.0   & 12.3     & 29.6     \\
                           & & RFS  & 22.7      & 37.3    & 23.9          & 9.2   & 21.3     & 30.0    \\
                           & & IRFS &\textbf{24.4}      & \textbf{39.8}   & \textbf{25.8}         & \textbf{14.1}  & \textbf{22.8}    & \textbf{30.7}    \\ \midrule
\multirow{3}{*}{Mask R-CNN} & \multirow{3}{*}{ResNet-101} & w/o sampling       & 18.6      & 30.0    & 19.6        & 0.0   & 14.2     & 31.6     \\
                           & & RFS  & 24.9      & 40.0    & 26.5          & 11.9   & 23.8     & 32.0    \\
                           & & IRFS & \textbf{26.4}      & \textbf{41.8}   & \textbf{28.3}         & \textbf{16.9}  & \textbf{24.7}    & \textbf{32.5}    \\ \midrule
\multirow{3}{*}{Cascade Mask R-CNN} & \multirow{3}{*}{ResNet-50} & w/o sampling       &  20.1     &  29.5   & 21.3         & 0.3   & 16.1     & 33.3    \\
                           & & RFS  & 26.6      & 38.5   & 28.1    & 12.7   & 25.7      & 33.7   \\
                           & & IRFS & \textbf{28.8}      & \textbf{41.4}   & \textbf{30.6}         & \textbf{19.1}  & \textbf{27.4}    & \textbf{34.4}    \\ \midrule
\multirow{3}{*}{Cascade Mask R-CNN} & \multirow{3}{*}{ResNet-101} & w/o sampling       & 21.8      & 31.5    &  23.1 &  0.2 &  18.4    & 35.0     \\
                           & & RFS  & 29.1     & 41.6    & 31.1   & 17.0   & 28.4     & 35.3     \\
                           & & IRFS & \textbf{30.0}      & \textbf{42.8}   & \textbf{32.1}         & \textbf{19.1}  & \textbf{29.0}    & \textbf{35.9}    \\ \midrule                           
                            \midrule
Framework                   & Backbone                   & \textbf{Segmentation}      & mAP$_{segm}$ & AP$_{50}$ & AP$_{75}$ & AP$_r$ & AP$_c$ & AP$_f$ \\ \midrule
\multirow{3}{*}{Mask R-CNN} & \multirow{3}{*}{ResNet-50} & w/o sampling       & 16.2      & 26.0    & 17.0          & 0.0   & 12.5     & 27.3     \\
                           & & RFS  & 21.9      & 34.8    & 23.2          & 10.4   & 21.1     & 27.8    \\
                           & & IRFS & \textbf{23.7}      & \textbf{37.1}   & \textbf{25.1}         & \textbf{14.9}  & \textbf{22.8}    & \textbf{28.5}    \\ \midrule
\multirow{3}{*}{Mask R-CNN} & \multirow{3}{*}{ResNet-101} & w/o sampling       & 17.6      & 27.9    &  18.7         & 0.0   &  14.2    &  29.0    \\
                           & & RFS  & 23.8      & 37.0    & 25.3          & 12.3   & 23.3     & 29.3    \\
                           & & IRFS & \textbf{25.3}      & \textbf{39.2}   & \textbf{26.9}         & \textbf{17.2}  & \textbf{24.3}    & \textbf{29.9}    \\ \midrule
\multirow{3}{*}{Cascade Mask R-CNN} & \multirow{3}{*}{ResNet-50} & w/o sampling       &  17.8     & 27.6    &  19.0       & 0.2   & 14.8  & 29.0     \\
                           & & RFS  & 23.8      & 36.3    & 25.4   & 12.3   & 23.3     & 29.4    \\
                           & & IRFS & \textbf{25.6}      & \textbf{38.8}   & \textbf{27.4}         & \textbf{17.3}  & \textbf{24.8}    & \textbf{30.1}    \\ \midrule
\multirow{3}{*}{Cascade Mask R-CNN} & \multirow{3}{*}{ResNet-101} & w/o sampling       &  19.3     & 29.5    & 20.6        & 0.2    & 16.8     & 30.6     \\
                           & & RFS  & 26.0      & 39.2    & 27.8         & 15.9   & 25.6     & 30.8    \\
                           & & IRFS & \textbf{26.8}      & \textbf{40.3}   & \textbf{28.5}         & \textbf{17.7}  & \textbf{26.3}    & \textbf{31.3}    \\ \midrule
\end{tabular}
\end{adjustbox}
\end{table}
\section{Experiments}
\subsection{Dataset}
In our experiments, we evaluate the proposed IRFS approach on the LVIS v1.0 dataset~\cite{lvis}. LVIS contains 1.3 million object instances across 120K images and includes 1203 categories. This dataset is known for its heavy long-tail distribution and has been categorized into three groups: \textit{frequent}, \textit{common}, and \textit{rare}, based on the frequency of occurrence. Rare categories are defined as those appearing in fewer than 10 images, common categories appear in more than 10 but less than 100 images, and frequent categories appear in over 100 images. The dataset is split into 100K training images and 20K validation images, and we perform our experiments on this split.

\subsection{Implementation Details}
We conducted all experiments using MMdetection~\cite{mmdetection}, a popular open-source library for object detection and instance segmentation. For our experiments, we utilized Mask R-CNN~\cite{8237584}, a state-of-the-art framework for object detection and instance segmentation. We trained Mask R-CNN with ResNet-50 and ResNet-101 backbones~\cite{DBLP:conf/cvpr/HeZRS16} and Feature Pyramid Network (FPN)~\cite{8099589}. We also performed experiments on ATSS~\cite{zhang2020bridging}, a one-stage detector. We followed the standard LVIS setup and hyperparameters for the models. Specifically, we trained the models using SGD with 0.9 momentum and a batch size of 16 on 4 GPUs (NVIDIA Tesla V100 with 32 GB VRAM), with an initial learning rate of 0.02 and weight decay of $10^{-4}$. The training and inference images were resized to a shorter and longer image edge of 800 and 1333 pixels, and we only used horizontal flipping for data augmentation unless stated otherwise. We trained all models for 12 epochs ($1\times$ schedule), with decay at the $8$th and $11$th epochs. We train the RFS models with the reported optimal value of $t=10^{-3}$. We use IRFS with geometric mean and $t=10^{-3}$ unless stated otherwise. In addition to RFS and IRFS, we report the performance of training without any sampling, i.e. $t=0$, and denote it in the Tables as ``w/o sampling''.
\subsection{Evaluation Metric}
We evaluated our models using the LVIS metrics. The metrics include mean average precision (mAP), average precision (AP) with intersection over union (IoU) of $50\%$ (AP$_{50}$), AP with IoU of $75\%$ (AP$_{75}$), AP on rare classes (AP$_r$), AP on common classes (AP$_c$), and AP on frequent classes (AP$_f$). For Mask R-CNN, we reported both detection and segmentation metrics, which are denoted as mAP$_{bbox}$ and mAP$_{segm}$, respectively.

\subsection{Main Results}
Table \ref{tab:table_mask_rcnns} illustrates the performance of IRFS with respect to counterpart RFS method. 

\textbf{Object Detection.} In all frameworks, training without a balanced sampling results in the lowest AP. Due to the heavy-tailed distribution, the w/o sampling approach suffers greatly for rare class detection. While RFS provides a significant improvement over the w/o sampling approach, the proposed IRFS achieves the best performance with a significant improvement over RFS across all metrics. In particular, for Mask R-CNN with a ResNet-50 backbone, our proposed IRFS provides a relative $+50\%$ improvement over RFS for rare categories. Moreover, the superior performance of IRFS for rare classes does not result in degraded results for common and frequent classes. In fact, it relatively improves the performance of common and frequent categories by $+7\%$ and $+2\%$, respectively, compared to RFS. For Mask R-CNN with a ResNet-101 backbone, IRFS outperforms the RFS by $5.0$ AP ($+40\%$) for rare categories and $1.5$ mAP ($+5\%$) for overall classes. We achieve similar consistent results with the more advanced Cascade Mask R-CNN~\cite{Cai_2018_CVPR} framework across ResNet-50 and ResNet-101 backbones.

\textbf{Instance Segmentation.} Similar to object detection, IRFS exhibits significant improvement over RFS in instance segmentation. Particularly, for Mask R-CNN with ResNet-50 backbone, IRFS achieves a mAP of $23.7\%$, resulting in a $+8\%$ improvement over RFS. For rare categories, IRFS achieves a $14.9\%$ AP, which is a $+43\%$ improvement over RFS. Consistently, IRFS outperforms RFS for common and frequent categories. With the ResNet-101 backbone and Cascade Mask R-CNN framework, IRFS maintains its superior performance compared to RFS across all cases.

\begin{figure}[t]
\centering
\includegraphics[width=\linewidth]{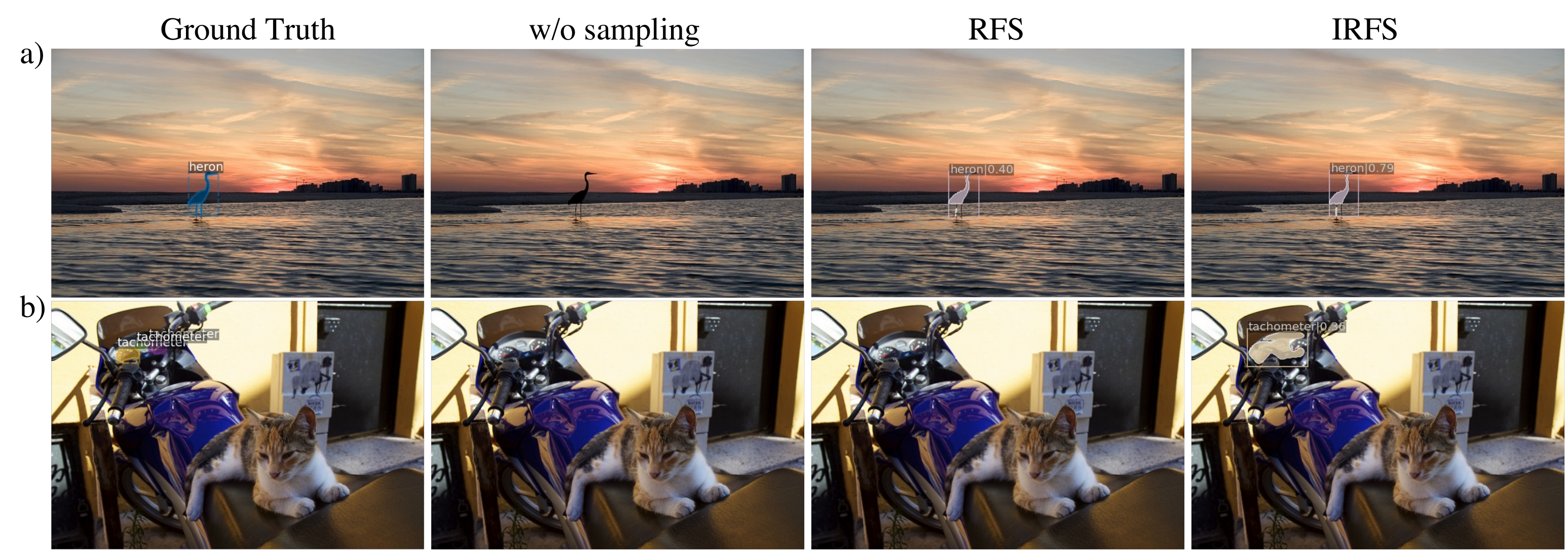}
\caption{Representative detection and instance segmentation visualizations on LVIS v1.0 dataset. For convenience, we present the results specifically focused on the rare categories.}
\label{fig:IRFS_main_res}
\end{figure}

Fig. \ref{fig:IRFS_main_res} illustrates the detection and segmentation results for w/o sampling, RFS and IRFS models trained with Mask R-CNN framework using ResNet-50 backbone on $1\times$ schedule. Note that only targeted rare categories are shown for visibility. As expected, w/o sampling fails in detecting the rare categories. IRFS detects and segment with a higher accuracy for rare categories detected by RFS as well (Fig \ref{fig:IRFS_main_res} a). IRFS further shows its strength by detecting rare categories that are not detected by RFS (Fig. \ref{fig:IRFS_main_res}b). 

\begin{table}[t]
\caption{Comparison of the combination of RFS and IRFS with ECM loss, using Mask R-CNN framework with a ResNet-50 backbone, on the LVIS v1.0 validation set.}
\label{tab:table_losses}
\begin{adjustbox}{width=.9\linewidth,center}
\begin{tabular}{lllll|llll}
\toprule
         & \multicolumn{4}{l|}{Detection} & \multicolumn{4}{l}{Segmentation} \\
\midrule
Method  & mAP$_{bbox}$   & AP$_r$ & AP$_c$ & AP$_f$   & mAP$_{segm}$    & AP$_r$ & AP$_c$ & AP$_f$   \\
\midrule
ECM    &       24.8 & 13.3      & 23.3    & 31.6     & 24.6      & 14.7    & 24.1  & 29.6      \\
ECM+RFS   &    26.8  & 16.5      & 26.2    & 31.9     & 26.4      & 18.7    & 26.2  & 30.0       \\
ECM+IRFS  &    \textbf{27.6}   & \textbf{19.4}      & \textbf{26.5}    & \textbf{32.4}     & \textbf{27.0}      & \textbf{20.6}    & \textbf{26.4}  & \textbf{30.6}     \\
\midrule
RFS      &  22.7       & 9.2       & 21.3      & 30.0      & 21.9       & 10.4       & 21.1   & 27.8  \\
\midrule
IRFS      &  24.4    & 14.1      & 22.8      & 30.7      & 23.7       & 14.9        & 22.8       & 28.5 
\end{tabular}
\end{adjustbox}
\end{table}

\begin{figure}[t]
\centering
\includegraphics[width=\linewidth]{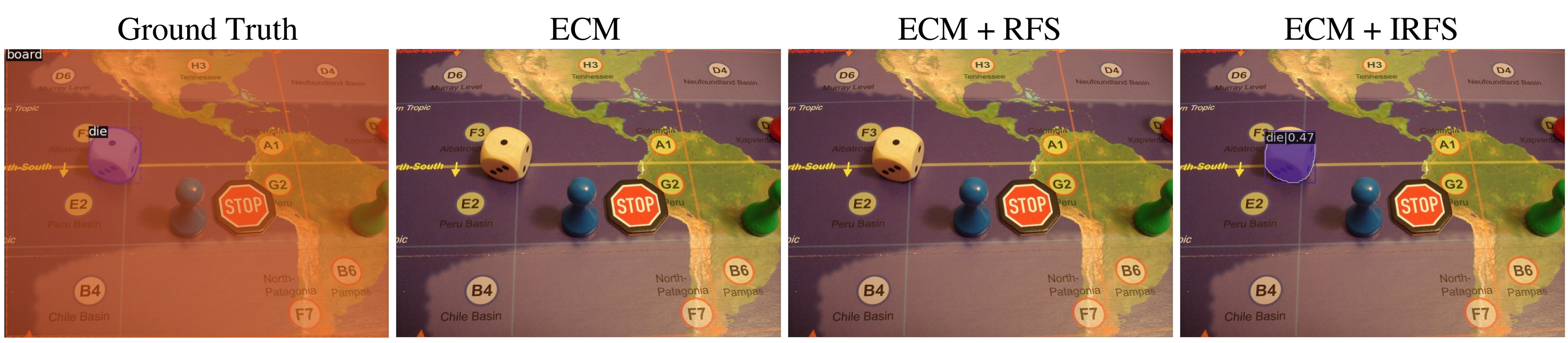}
\caption{Illustration  of detection and instance segmentation results for ECM re-weighting method and its combination with re-sampling approaches. For convenience, we exclusively display the results for the targeted rare categories.}
\label{fig:IRFS_ECM_res}
\end{figure}

\begin{table}[!t]
\caption{Comparison of RFS and IRFS using a single-stage detector ATSS with different backbones on the LVIS v1.0 validation set.}
\label{tab:table_single_stage}
\begin{adjustbox}{width=.8\linewidth,center}
\begin{tabular}{lllllll}
\toprule
Framework             & Backbone                    & Method & mAP$_{bbox}$ & AP$_r$ & AP$_c$ & AP$_f$ \\
\midrule
\multirow{2}{*}{ATSS} & \multirow{2}{*}{ResNet-50}  & RFS    &   19.3  & 7.5    & 17.2    & 26.7    \\
                      &                             & IRFS   &    \textbf{19.7} &  \textbf{9.5}   & \textbf{17.2}    & \textbf{26.9}    \\
\midrule
\multirow{2}{*}{ATSS} & \multirow{2}{*}{ResNet-101} & RFS    & 20.9    & 8.8    & 19.0    & 28.4    \\
                      &                             & IRFS   & \textbf{22.0}    & \textbf{10.7}    & \textbf{20.4}    & \textbf{28.6}   
\end{tabular}
\end{adjustbox}
\end{table}

Table \ref{tab:table_losses} presents a comparison of the impact of class and instance balanced sampling approaches on the state-of-the-art weighted loss function. It is observed that ECM without any class balanced sampling achieves a better overall bounding box and mask AP compared to RFS and IRFS. However, IRFS shows better results in terms of the target rare categories compared to standalone ECM. Class/Instance balanced sampling and weighted loss are complementary, and ECM with RFS and IRFS achieves significantly improved performance compared to standalone ECM or RFS/IRFS. Specifically, ECM with the proposed IRFS achieves the best performance across all metrics. For target rare categories, it achieves a $19.4\%$ and $20.6\%$ AP for detection and segmentation tasks, respectively, which is a relative $17\%$ and $10\%$ improvement compared with ECM with RFS for detection and segmentation tasks.

Fig. \ref{fig:IRFS_ECM_res} illustrates the detection and segmentation results for ECM, ECM $+$ RFS and ECM $+$ IRFS models trained with Mask R-CNN framework using ResNet-50 backbone on $1\times$ schedule. Note that only targeted rare categories are shown for visibility. In this challenging representative scenario, ECM without any re-sampling as well as RFS and IRFS without ECM fails to detect the rare categories. While ECM combined with RFS also fails detecting any rare category, ECM with IRFS achieves a better performance and successfully detects and segments a rare category.

Table \ref{tab:table_single_stage} presents the performance comparison of RFS and IRFS on ATSS, a state-of-the-art single-stage detector, trained with ResNet-50 and ResNet-101 backbones. When trained with IRFS, ATSS exhibits a substantial relative improvement of $26\%$ and $21\%$ over its RFS-trained counterpart on rare classes, using ResNet-50 and ResNet-101 backbones, respectively. Furthermore, it consistently achieves better overall mAP compared to the counterpart RFS method.

\begin{table}[t]
\caption{An ablation study on IRFS with different thresholds $t$ using Mask R-CNN with a ResNet-50 backbone on a $1\times$ schedule on LVIS v1.0 validation set.}
\label{tab:table_thresholds}
\begin{adjustbox}{width=.9\linewidth,center}
\begin{tabular}{lllll|llll}
\toprule
         & \multicolumn{4}{l|}{Detection} & \multicolumn{4}{l}{Segmentation} \\
\midrule
$t$  & mAP$_{bbox}$   & AP$_r$ & AP$_c$ & AP$_f$   & mAP$_{segm}$    & AP$_r$ & AP$_c$ & AP$_f$   \\
\midrule
w/o sampling ($t=0$)      & 16.9      &  0.0     & 12.3      & 29.6       & 16.2       & 0.0       & 12.5       & 27.3      \\
\midrule
$t=0.1$    &       19.8&       6.0&       17.8& 28.0       &18.9    &6.0        & 17.5        & 26.1       \\
$t=0.01$   &  23.7     & 11.0      & 22.2      & \textbf{30.9}      & 22.9       & 11.3       &  22.2      & \textbf{28.7}      \\
$t=0.001$  &  \textbf{24.4}     & \textbf{14.1}      & \textbf{22.8}      & 30.7      & \textbf{23.7}       & \textbf{14.9}        & \textbf{22.8}       & 28.5      \\
$t=0.0001$ &  19.5     & 3.7      &  16.5     & 29.7     & 18.7       &  3.9      & 16.8       & 27.4      \\ 
\midrule
RFS ($t=0.001$)      & 22.7       & 9.2       & 21.3      & 30.0      & 21.9       & 10.4       & 21.1   & 27.8    
\end{tabular}
\end{adjustbox}
\end{table}

\begin{table}[t]
\caption{An ablation study on different averaging options for IRFS using Mask R-CNN with a ResNet-50  backbone on a $1\times$ schedule on LVIS v1.0 validation set. We also propose instance only approach  for comparison purposes.}
\label{tab:table_averages}
\begin{adjustbox}{width=.9\linewidth,center}
\begin{tabular}{lllll|llll}
\toprule
         & \multicolumn{4}{l|}{Detection} & \multicolumn{4}{l}{Segmentation} \\
\midrule
Mean  & mAP$_{bbox}$   & AP$_r$ & AP$_c$ & AP$_f$   & mAP$_{segm}$    & AP$_r$ & AP$_c$ & AP$_f$   \\
\midrule
w/o sampling ($t=0$)     & 16.9      &  0.0     & 12.3      & 29.6       & 16.2       & 0.0       & 12.5       & 27.3      \\
\midrule
Geometric    &  24.4     & 14.1      & 22.8      & 30.7      & 23.7       & 14.9        & 22.8       & 28.5      \\
Harmonic   & 24.5         & \textbf{15.0}       & 22.7      & 30.8      & \textbf{23.8}       & \textbf{15.5}       & 22.7   & 28.6      \\
Arithmetic  & 24.1         & 12.2       & \textbf{23.2}      & 30.5      & 23.3       &  12.9     & \textbf{23.0}   &  28.3     \\
Quadratic &  23.6      &  11.7      & 22.2      & 30.4      & 22.7       & 12.0      & 22.0   & 28.2      \\ \midrule
Instance Only & \textbf{24.6}       & 14.4       & 22.9      & \textbf{31.0}     & 23.7     & 14.0     & 22.9  & \textbf{28.7}    \\
\midrule
RFS      &  22.7       & 9.2       & 21.3      & 30.0      & 21.9       & 10.4       & 21.1   & 27.8    
\end{tabular}
\end{adjustbox}
\end{table}

\subsection{Ablation Study}
In this section we conduct ablation studies on two key parameters of the proposed IRFS method, namely threshold value and averaging method.

Table \ref{tab:table_thresholds} presents the ablation study for IRFS using thresholds, $t \in \{10^{-1}, 10^{-2}, 10^{-3}, 10^{-4}\}$. The best overall mAP is achieved by IRFS with $t=10^{-3}$. For the target rare category, $t=10^{-3}$ significantly outperforms other thresholds for both detection and segmentation. It is noteworthy that IRFS's best performing threshold value, $t=10^{-3}$ matches RFS's best matching threshold value \cite{lvis}.

Table \ref{tab:table_averages} shows the results of different common averaging options, including geometric, harmonic, arithmetic, and quadratic averaging.\footnote{Note the mathematical definition for geometric, harmonic, arithmetic, and quadratic average for two numbers $x$ and $y$ is defined as $\sqrt{xy}$, $\frac{2xy}{x+y}$, $\frac{x+y}{2}$, and $\sqrt{\frac{x^2+y^2}{2}}$, respectively.} In addition to averaging, we also propose and include the instance only repeat factor sampling. Instance only disregards the image count in the re-sampling process. We present the results for $t=10^{-3}$. Both harmonic and geometric averaging show improved performance over arithmetic and quadratic averaging, with the former providing the best overall results. The results indicate that all averaging options provide significant improvement over RFS. This further emphasizes the importance of including instances when addressing long-tail distributions. More interestingly, instance only approach achieves the best overall mAP in detection and it achieves more than $+50\%$
improvement over RFS, the counterpart image count only based re-sampling method.

\section{Discussion and Conclusions}
In this work, we presented a novel approach named instance-aware repeat factor sampling (IRFS) for handling long-tail distributions in object detection. Extensive experimental results demonstrate that by incorporating instance count in addition to image count, IRFS consistently outperforms RFS by significantly improving detection performance of rare categories while maintaining the overall performance for all categories. These results emphasize the importance of instance count in improving the performance of rare categories. As a matter of fact, a re-sampling approach based solely on instance count significantly outperforms the approach based solely on image count. However, disregarding image count can lead to suboptimal results when dealing with real-world long-tailed datasets that exhibit variations in image instance count distributions. Thus, we assert that IRFS is better suited for addressing long-tailed datasets with diverse distributions. 

IRFS is a simple method that can be used as a plug-in replacement for RFS, or combined with re-weighting methods to achieve state-of-the-art results in long-tailed object detection and instance segmentation. Due to its simplicity, easy integration with other methods, and effectiveness in improving performance, we believe IRFS can become a strong baseline for methods aiming to address long-tail distributions. For future work, we plan to investigate the application of IRFS in related fields in the computer vision domain, and explore new approaches to optimize the averaging method.
\bibliographystyle{named.bst}
\bibliography{neurips_2023}

\end{document}